\documentclass[10pt,twocolumn,letterpaper]{article}

\usepackage{cvpr}              

\usepackage{graphicx}
\usepackage{amsmath}
\usepackage{amssymb}
\usepackage{booktabs}
\usepackage{float}
\usepackage{widetable}
\usepackage[accsupp]{axessibility} 

\usepackage[pagebackref,breaklinks,colorlinks]{hyperref}

\usepackage[capitalize]{cleveref}
\crefname{section}{Sec.}{Secs.}
\Crefname{section}{Section}{Sections}
\Crefname{table}{Table}{Tables}
\crefname{table}{Tab.}{Tabs.}

\def\cvprPaperID{113} 
\def\confName{CVPR}
\def\confYear{2022}

\begin{document}

\title{Efficient Progressive High Dynamic Range Image Restoration via Attention and Alignment Network}

\author{Gaocheng Yu\footnotemark[2] \and Jin Zhang\footnotemark[2] 
\and
Zhe Ma\and Hongbin Wang \\
AntGroup\\
{\tt\small \{yugaocheng.ygc, zj346862, mz281827, hongbin.whb\}@antgroup.com}
}

\twocolumn[{%
\maketitle

\begin{figure}[H]
\hsize=\textwidth
\centering
\begin{subfigure}{0.23\textwidth}
  \includegraphics[width=\textwidth]{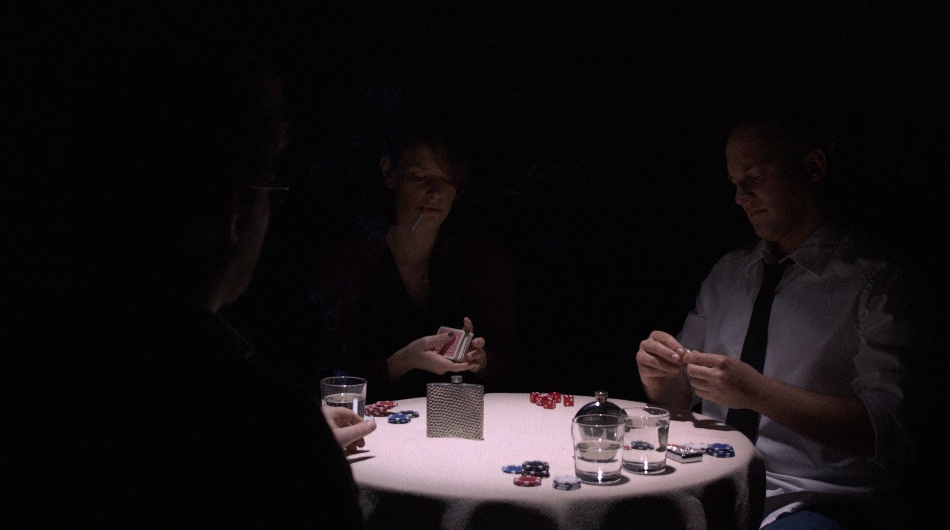}
  \vfill
  \includegraphics[width=\textwidth]{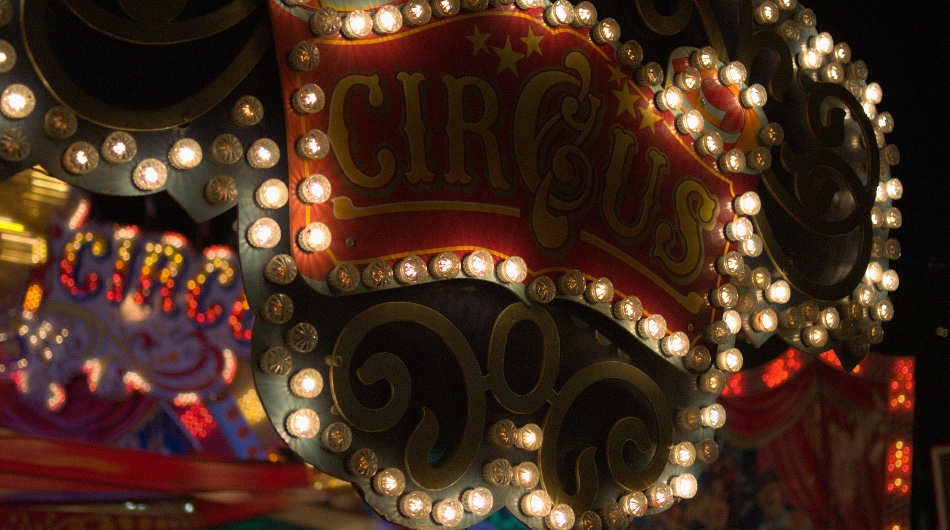}
  \vfill
  \includegraphics[width=\textwidth]{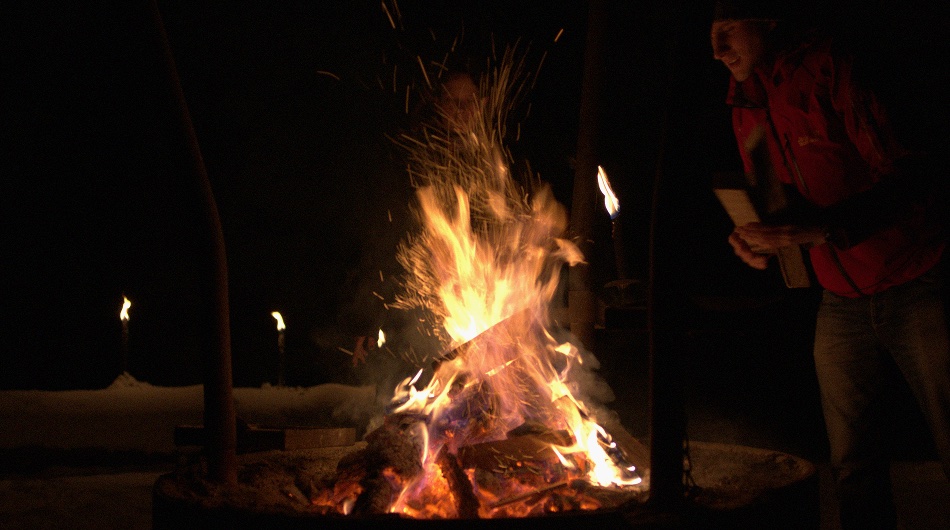}
  \vfill
  \includegraphics[width=\textwidth]{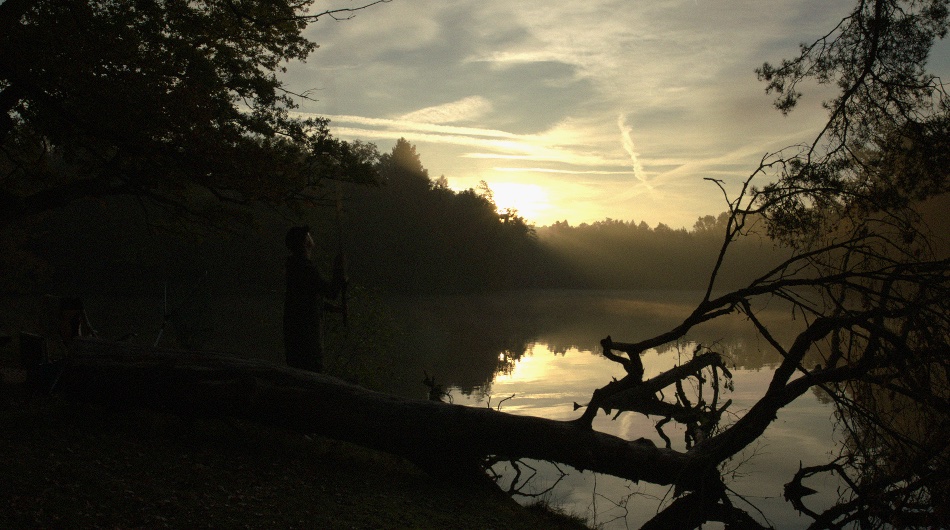}
  \caption{LDR short exposed.}
  \label{fig:vis_a}
\end{subfigure}
\hfill
\begin{subfigure}{0.23\textwidth}
  \includegraphics[width=\textwidth]{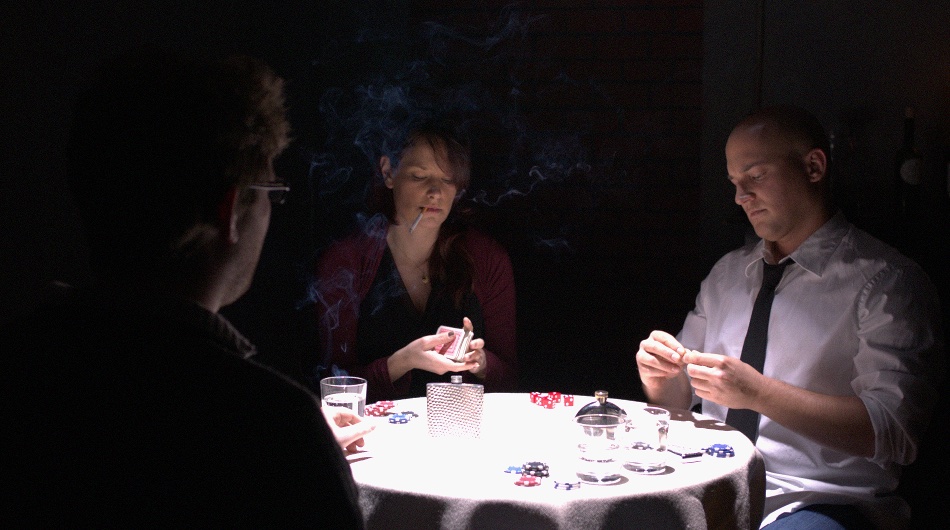}
  \vfill
  \includegraphics[width=\textwidth]{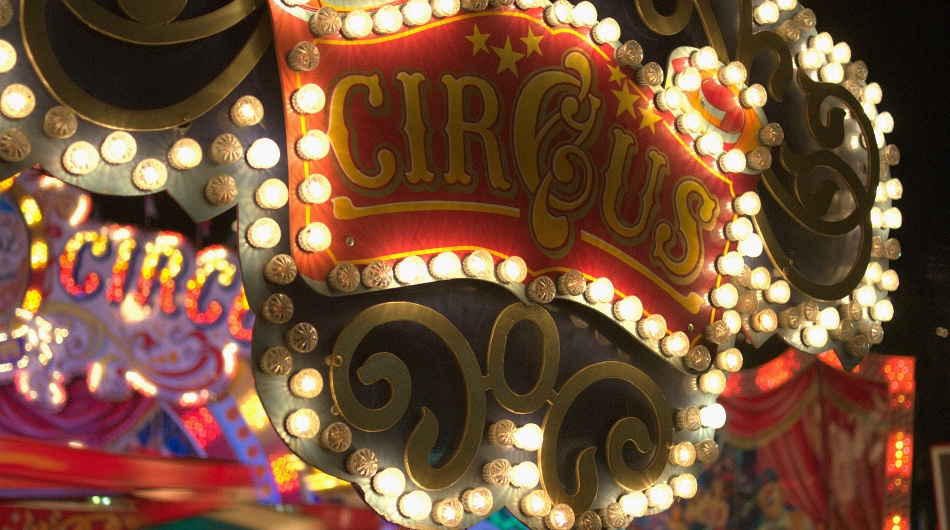}
  \vfill
  \includegraphics[width=\textwidth]{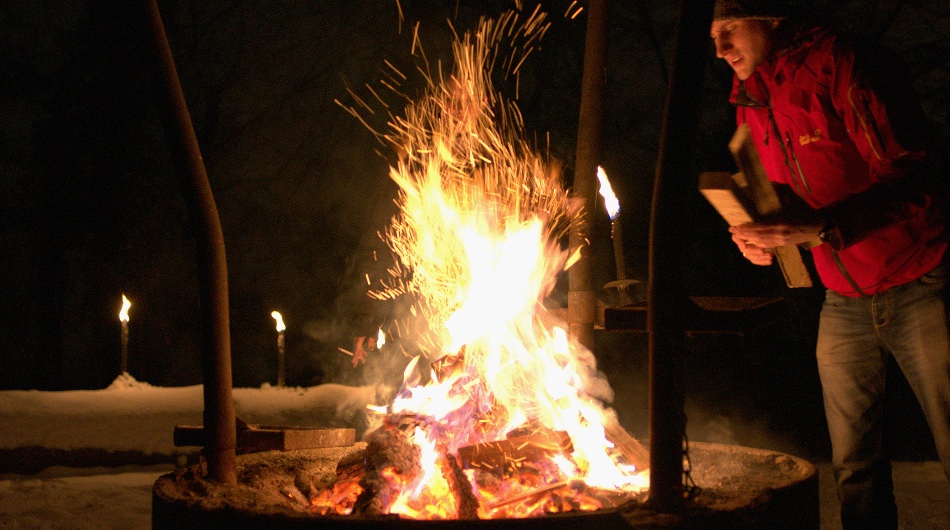}
  \vfill
  \includegraphics[width=\textwidth]{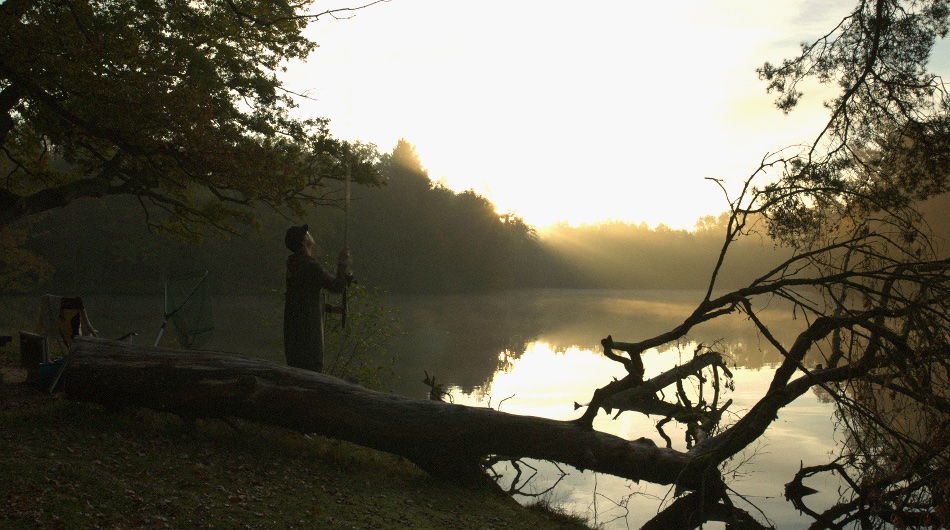}
  \caption{LDR medium exposed.}
  \label{fig:vis_b}
\end{subfigure}
\hfill
\begin{subfigure}{0.23\textwidth}
  \includegraphics[width=\textwidth]{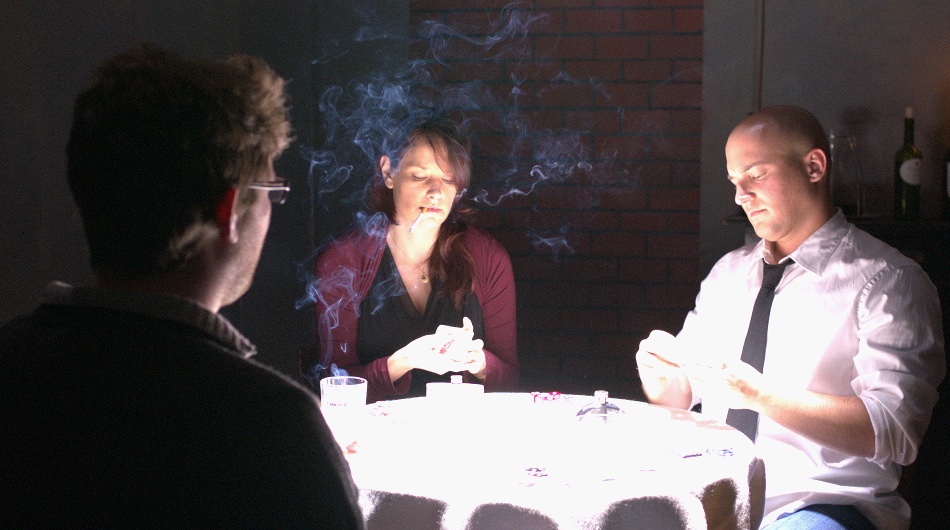}
  \vfill
  \includegraphics[width=\textwidth]{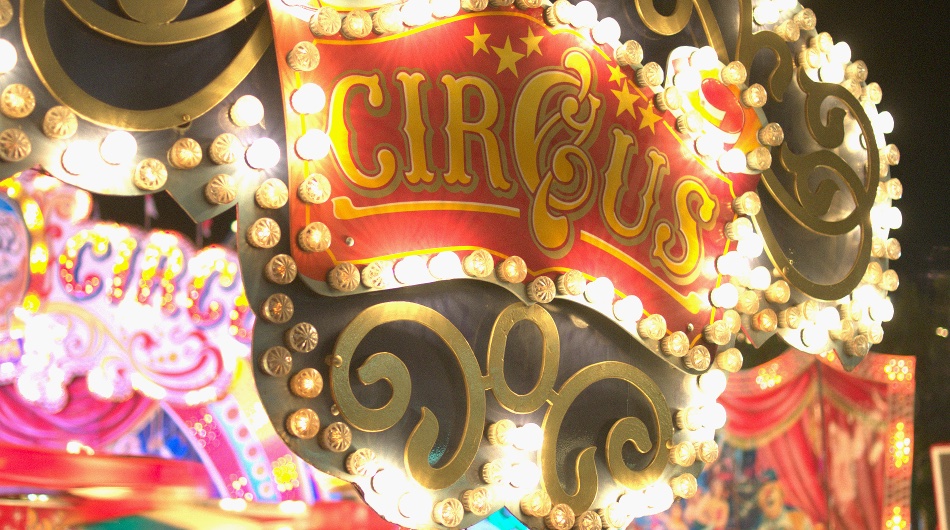}
  \vfill
  \includegraphics[width=\textwidth]{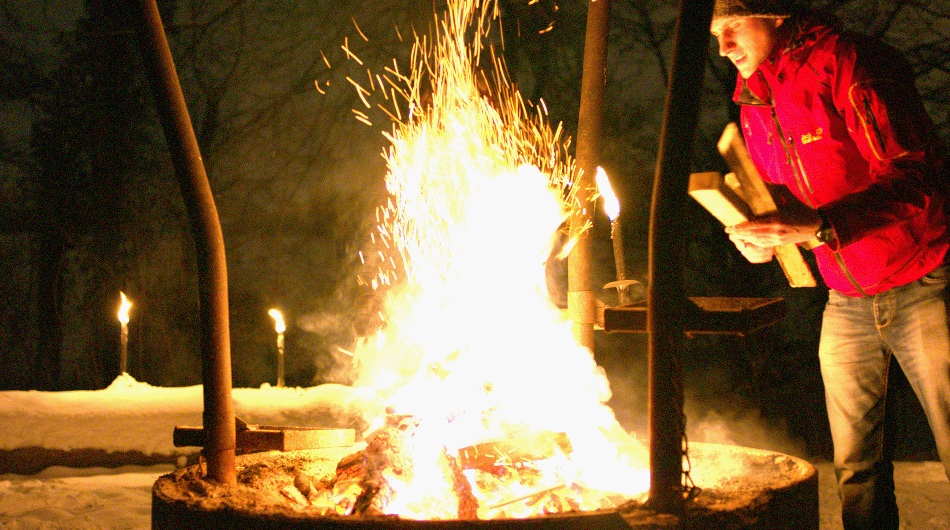}
  \vfill
  \includegraphics[width=\textwidth]{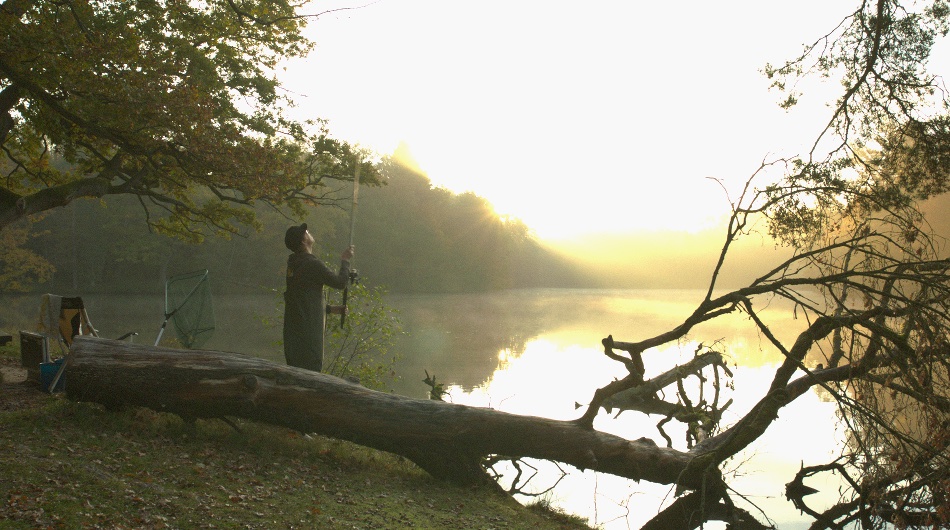}
  \caption{LDR long exposed.}
  \label{fig:vis_c}
\end{subfigure}
\hfill
\begin{subfigure}{0.23\textwidth}
  \includegraphics[width=\textwidth]{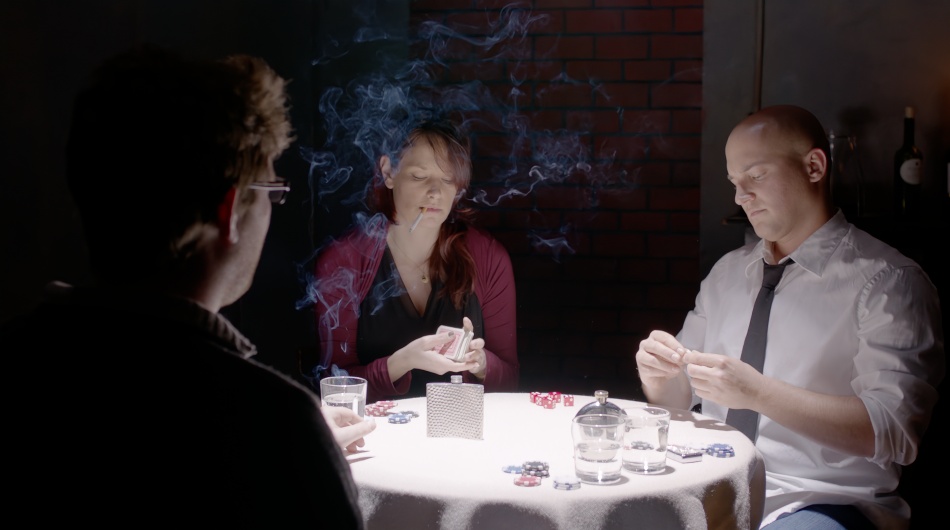}
  \vfill
  \includegraphics[width=\textwidth]{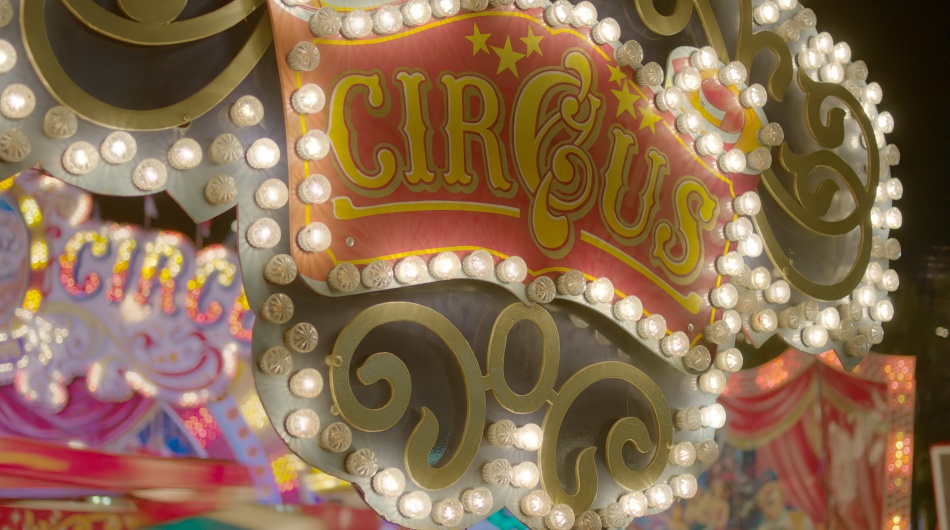}
  \vfill
  \includegraphics[width=\textwidth]{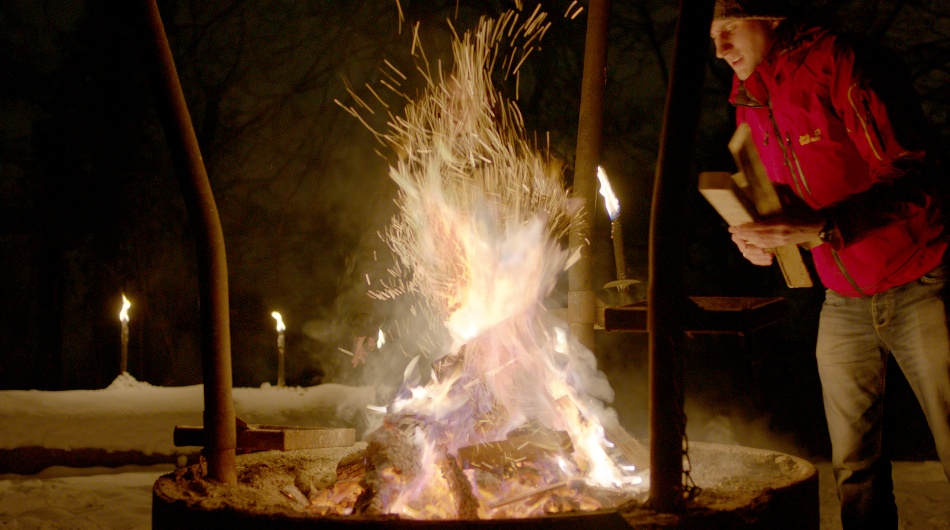}
  \vfill
  \includegraphics[width=\textwidth]{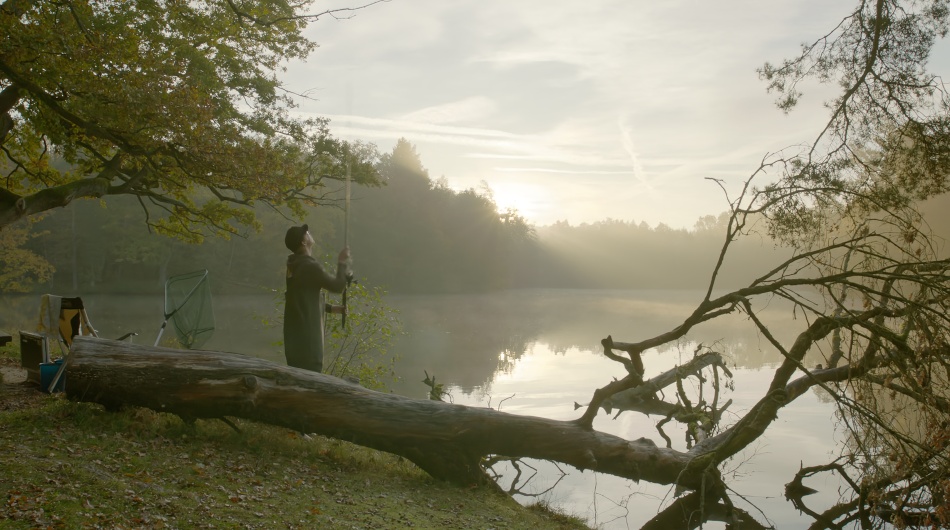}
  \caption{HDR reconstructed by EAPNet.}
  \label{fig:vis_d}
\end{subfigure}

\caption{Examples of HDRs restored from LDRs by EAPNet.
\cref{fig:vis_a}, \cref{fig:vis_b} and \cref{fig:vis_c} are short, medium, and long exposed LDRs respectively,
and the HDR images restored by our standard EAPNet are presented in \cref{fig:vis_d}.}
\label{fig:visualize}

\end{figure}

}]

\renewcommand{\thefootnote}{\fnsymbol{footnote}} 
\footnotetext[2]{These authors contributed equally to this work.} 

\begin{abstract}
HDR is an important part of computational photography technology. 
In this paper, we propose a lightweight neural network called \textbf{Efficient Attention-and-alignment-guided Progressive Network (EAPNet)} for the challenge NTIRE 2022 HDR Track 1 and Track 2. 
We introduce a multi-scale lightweight encoding module to extract features. 
Besides, we propose Progressive Dilated U-shape Block (PDUB) which is a progressive plug-and-play module for dynamically tuning MAccs and PSNR. 
Finally, we use fast and low-power feature-alignment module to deal with misalignment problem in place of the time-consuming Deformable Convolutional Network (DCN). 
The experiments show that our method achieves about 20$\times$ compression on MAccs with better PSNR-$\mu$ and PSNR compared to the state-of-the-art method. 
We got the 2$^{nd}$ place of both two tracks during the testing phase. 
\cref{fig:visualize} shows the visualized result of NTIRE 2022 HDR challenge.
\end{abstract}

\section{Introduction}
\label{sec:intro}

HDR image restoration task aims to recover image contents, details, and color from one or several LDRs. Due to the limitations of camera sensor, the LDRs degraded model can be described as in \cite{perezpellitero22}: 

\begin{equation}
  I_f = min\{\varPhi t/g + I_0 + n, I_{max}\}
  \label{eq:model}
\end{equation}
where $I_f$ denotes LDR image observations, $\varPhi$ is the scene brightness, $t$ is the exposure time, $g$ is the sensor gain, $I_0$ is the constant offset current, $n$ is the sensor noise, and $I_{max}$ denotes the saturation point\cite{perezpellitero22}.

In the recent, learnable approaches attempt to recover HDR images via convolutional neural networks (CNNs)\cite{HDRUnet, related_eilertsen2017hdr}. There are two ways of HDR image restoration: single frame restoration and multi frame restoration. The former focuses on learning a mapping from LDR to HDR\cite{Zero-DCE,HDRUnet}, and the latter processes Multi-Exposure Fusion (MEF), and works on the misalignment problem \cite{FlexHDR} additionally. 

Generally, the ISP pipeline\cite{related_hasinoff2016burst, introcution_AG, introcution_pixelv4} shows that the LDRs suffer from problems such as motions, noises, truncations, and saturations in the image signal process. Taking the above issues into account, researchers work on optical flow\cite{introcution_raft} and denoise\cite{introcution_dcnn} technologies jointly with HDR. Meanwhile, the application of HDR restoration technology requires real time performance on mobile devices\cite{HDRNet}. Due to the highly resolution of input LDRs, low-power and efficient technology will become more challenging than image detection\cite{introcution_coco} and image classification\cite{introcution_imagenet} tasks.



In this paper, we propose a lightweight network Efficient Attention-and-alignment-guided Progressive Network (EAPNet) for HDR restoration task. We consider the balance between performance and efficiency, and adopt depthwise separable convolution\cite{related_mbnv1,related_mbnv2,related_mbnv3}. To deal with misalignment problem, we introduce the feature-alignment network\cite{feature_align} to predict scale and bias to describe the local offset instead of DCN. And we propose a new module called Progressive Dilated U-shape Block (PDUB) with low computation cost and good performance. We participated in NTIRE 2022 High Dynamic Range Challenge (both two tracks), and won the 2nd place in the two tracks. 

Our main contributions are summarized as follow:
\begin{itemize}

\setlength{\itemsep}{0pt}

\item Compared with the SOTA method\cite{yan2019attention}, we achieve a maximum 20x MAccs's compression (NTIRE 2022 High Dynamic Range Challenge, Track 2) under the same PSNR and PSNR-$\mu$.
\item We propose Progressive Dilated U-shape Block (PDUB), and the experiments show that we achieve better performance compared with DRDB\cite{yan2019attention}.
\item We introduce a lightweight Feature-Alignment module to deal with misalignment problem with low computation cost.
\item Taking noise and MAccs into consideration, we propose an efficient Multi-Scale Encoder layer.
\end{itemize}

\begin{figure*}
  \centering
   \includegraphics[width=0.9\linewidth]{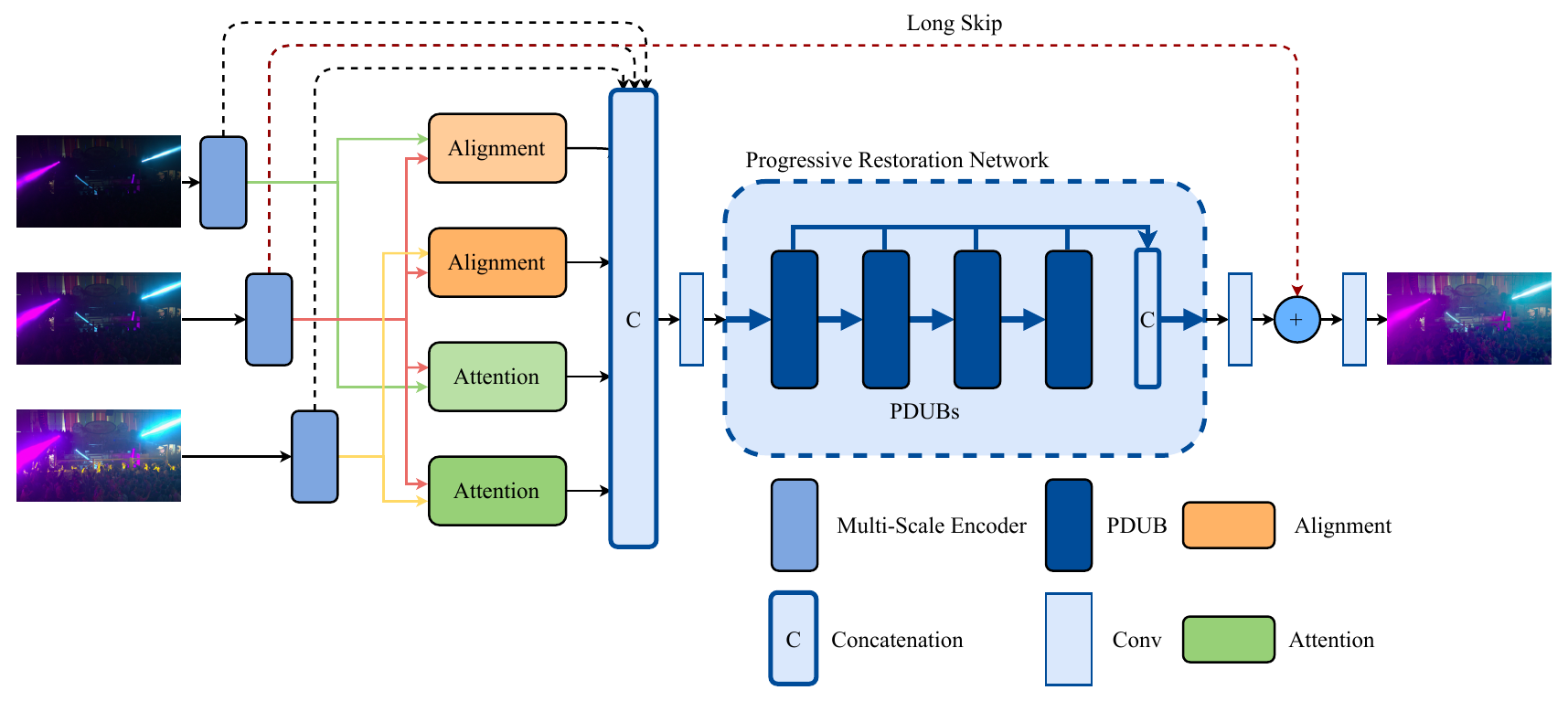}
   \caption{Overall structure of EAPNet.}
   \label{fig:overall}
\end{figure*}

Figure~\ref{fig:overall} shows the overall structure of our method.

\section{Related Work}
\subsection{Single Frame HDR}
Single Frame HDR has been extensively studied. Typically, without the negative impact of ghost artefact, single frame HDR researches on the LDR-to-HDR mapping problem. \cite{related_eilertsen2017hdr, related_yang2018image} propose a CNN based method to predict the HDR images directly. \cite{HDRUnet, related_zheng2021ultra} use a network to predict bilateral grid of coefficients in low-resolution, and lead to image enhancement. \cite{related_marnerides2018expandnet} devises a multi-branch network which extracts the local and global features respectively, and fuses them via concatenation operator, leading to a good performance.

Researches on efficient and low-power single frame HDR solutions are wide. \cite{related_a2021two,HDRUnet} propose a dequantize network to recover missing details from the low-bit LDR image. And \cite{Zero-DCE, related_li2021learning} formulate the HDR task as a curve estimation problem via neural network, and adjust the LDRs pixel-wise. All of the above methods learn a compressing space to describe the mapping between LDR and HDR, discarding the details of under or over exposed area.
\subsection{Multi Frame HDR}

Specifically, with the bracketed exposure strategy, \cite{related_mertens2007exposure, related_hasinoff2016burst} use image pyramid, and calculate brightness weight, contrast weight, and exposure weight to fuse LDRs. Furthermore, \cite{related_ma2017robust, related_li2020fast} propose a Structural Patch Decomposition (SPD) approach, decompose the signal into intensity and structure components, restore them individually, and fuse the restored components. The traditional approaches generally recover HDR image by means of decomposition and reconstruction.

Many learnable methods have been developed\cite{HDRUnet, ADNet, FlexHDR, relted_wu2018deep}. \cite{relted_wu2018deep} proposes a merge and fine-tune architecture to solve large-scale motions. \cite{HDRUnet, ADNet,FlexHDR} adopt an attention-guided mechanism to accomplish ghost-free merging. \cite{ADNet} additionally uses Deformable Convolution Network (DCN) to predict optical flow and align the other frames to the reference frame, achieving a good performance on detail restoration.

\subsection{Efficient Convolutional Neural Network}
The MobileNets\cite{related_mbnv1, related_mbnv2, related_mbnv3} decompose the standard convolution operator into two parts: depthwise convolution and pointwise convolution. \cite{related_ghostnet} uses group convolution to reuse redundant features. \cite{related_Shufflenetv1,related_Shufflenetv2} propose a channel-shuffle operator, receiving a high precision on image classification and detection tasks. \cite{related_LKA} introduces the attention mechanism on the basis of \cite{related_mbnv1, related_mbnv2, related_mbnv3}, and proposes a Large Kernel Attention (LKA) module to enable self-attention and self-adaption according to the high-wide receptive field.

\section{Proposed Method}
\label{sec:prop}
We propose an end-to-end network EAPNet consisting of a Feature Extraction Network and a Progressive Restoration Network. 
The overall structure of EAPNet is presented in \cref{fig:overall}.

The Feature Extraction Network first extracts features with a Multi-Scale Encoder Module, and then attends features with an Attention Module and a Feature-Alignment Module.
The Processive Restoration Network first fuses features by guidance from attention and alignment module,
and then progressively restores features with proposed PDUBs.
The restored features are then upsampled to the original resolution to fuse with the long skip feature, and then recovered with several standard convolutional layers.

We consider three LDR images, i.e., $I_i, i = 1, 2, 3$ as input and let the second LDR image $I_2$ be the reference frame.
With the Exposure Value (EV), we conduct EV alignment, and directly concatenate them with the original input, and get 6-channel input:
\begin{equation}
    I^{input}_i = concat([I_i, f(I_i)]), \quad i = 1, 2, 3
    \label{eq:base}
\end{equation}
where $I^{input}_i$ is the input of each frame, $I_i$ is the original LDR image, and $f$ is the EV alignment function.

\textbf{Depthwise Separable Convolution}.
Depthwise separable convolution, proposed by \cite{related_mbnv1}, factorizes a standard convolution into a depthwise convolution and a $1 \times 1$ convolution called pointwise convolution.
The depthwise separable convolution can drastically reduce computation cost and model size.
We replace the standard convolution in AHDR with depthwise separable convolution, except for the first one in the encoder and the last one in the restoration network.

\textbf{Multi-Scale Encoder Module}.
\begin{figure}[t]
  \centering
   \includegraphics[width=0.9\linewidth]{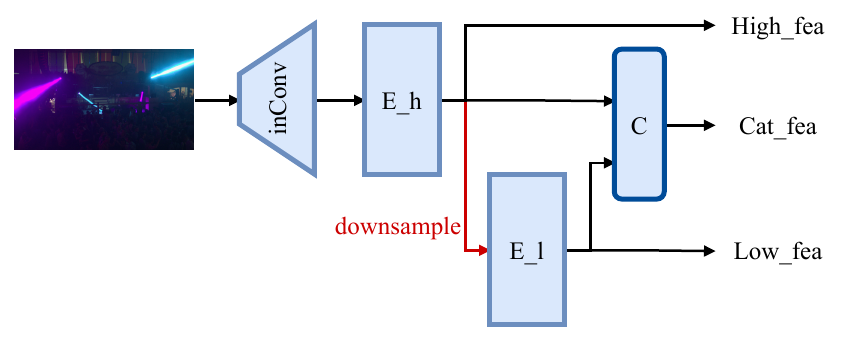}
   \caption{Structure of Multi-Scale Encoder.}
   \label{fig:msenc}
\end{figure}
Given the LDRs, we propose the Multi-scale Encoder module (MSEnc) to encode different scales of information with different frequencies.
As shown in \cref{fig:msenc}, the MSEnc first encodes the input with a standard convolution layer, and then extracts high resolution features which usually contain high frequency features with a high extractor.
The high features are then downsampled and processed by a low extractor to produce low resolution features.
The high and low features are further processed by attention and alignment module separately, and concatenated features are passed to the restoration network.
The reference frame features are also passed to the restoration network as the long skip feature.

\begin{equation}
  f_{h,i} = E_{h,i}(I^{input}_i), \quad i = 1, 2, 3
  \label{eq:ms_h}
\end{equation}
\begin{equation} 
  f_{l,i} = E_{l,i}(f_{h,i}), \quad i = 1, 2, 3
  \label{eq:ms_l}
\end{equation}
\begin{equation}
  f_i = concat([f_{h,i}, f_{l,i}]), \quad i = 1, 2, 3
  \label{eq:ms_ref}
\end{equation}
where $E_{h,i}$ and $E_{l,i}$ are $i_{th}$ high and low convolutional encoders, with a stride of 1 and 2 separately, and $f_{h,i}$ and $f_{l,i}$ are high and low features, and $f_i$ are concatenated features output from MSEnc.
We set the convolution channel to be 32, and obtain 64 channel features by concatenating multi-scale features, which is more efficient and effective than naive 64-channel convolutional encoders.

\begin{figure}[t]
  \centering
   \includegraphics[width=0.9\linewidth]{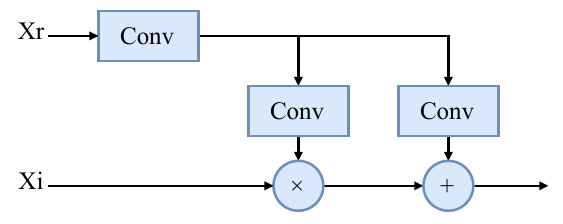}
   \caption{Structure of Feature-Alignment Module.}
   \label{fig:align}
\end{figure}

\textbf{Feature Alignment Module}.
We introduce a Feature-Alignment Module\cite{young2022feature} shown in \cref{fig:align}, to deal with misalignment occurred during camera capture.
Specifically, we concatenate the reference frame and input frame, providing spatially variant manipulations on half resolution.
While in the lightweight model, we keep full resolution for performance balance.

\textbf{Attention Module}.
Following \cite{yan2019attention}, we introduce the attention module to guide feature merging of the short and long exposed image with the reference image.
Taking the encoded feature as inputs, the attention module first concatenates the two input features, 
and then a convolutional layer followed by a sigmoid activation is applied to obtain the attention map.
The predicted attention map is finally used to guide feature merging by pointwise multiplication.

\begin{figure}[t]
  \centering
   \includegraphics[width=0.9\linewidth]{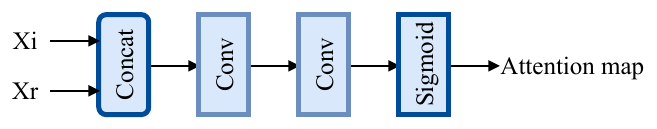}
   \caption{Structure of Attention Module.}
   \label{fig:attention}
\end{figure}

\begin{figure}[t]
  \centering
   \includegraphics[width=0.9\linewidth]{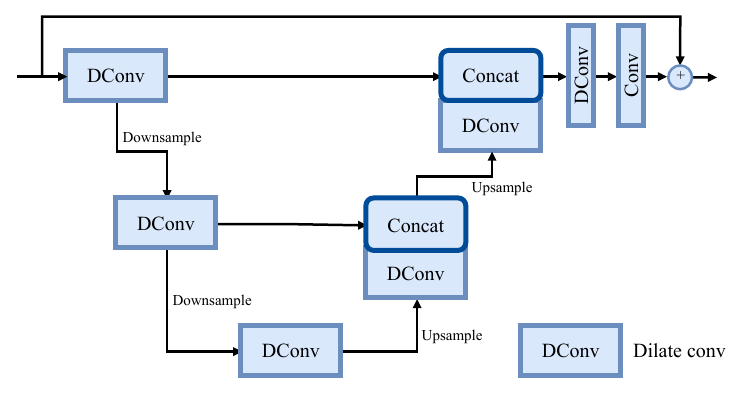}
   \caption{Structure of PDUB.}
   \label{fig:pdub}
\end{figure}

\textbf{Progressive Dilated U-shape Block}.
One of the main problems of high dynamic reconstruction is to solve the local and global fusion problem.
We propose Progressive Dilated U-shape Block (PDUB) to expand the receptive fields to better guide local feature reconstruction.
As shown in \cref{fig:pdub}, the PDUB is a tiny-UNet which consists 3 dilated encoder and decoders.
Since PDUB works at a smaller resolution, we can obtain a good balance between PSNR and MAccs.

The Progressive Restoration Network in our standard model consists of four consecutive PDUBs.
While in the lightweight model, the Progressive Restoration Network is shortened by cutting one PDUB, and performance is maintained through transposed convolution and strided convolution downsampling.

\textbf{Loss}. We also study the influences of different losses. Generally, optimizing the network in the tonemapped domain is better than directly in the signal space according to \cite{ADNet, relted_wu2018deep}. $\mu$-law $L_1$ loss measures $L_1$ loss in tangential and tonemapped space with $99$ percentile normalization, while \cite{HDRUnet} proposes $Tan\_L_1$ loss which measures $L_1$ loss in tangential space, enhancing low luminance values.

 To handle relative and absolute relationships between pixel values in HDR space due to normalize operator, we combine both two losses described in the following formula:
\begin{equation}
  \varGamma = \alpha * \mathcal{L} + \beta * \mathcal{G}
   \label{eq: our-loss}
\end{equation}

\begin{equation}
  \mathcal{T} (x) = \frac{log(1+\mu{x})}{1 + \mu} 
  \label{eq: ulaw}
\end{equation}

\begin{equation}
  \mathcal{L} = \left\lVert \mathcal{T}(I^{GT}) - \mathcal{T}(I^{H})\right\rVert_1 
  \label{eq: ulaw-l1}
\end{equation}

\begin{equation}
  \mathcal{G} = \left\lVert Tanh(I^{GT}) - Tanh(I^{H})\right\rVert_1
   \label{eq: tanh-l1}
 \end{equation}
where $\varGamma$ denotes our proposed loss function, $\mathcal{T} ( \cdot )$ denotes tonemapping operator, $\mathcal{L}$ denotes $\mu$-law $L_1$ loss, and $\mathcal{G}$ denotes $Tan\_L_1$ loss. 
We set $\mu$ to $5000$, and $\alpha$, $\beta$ to $0.5$ in this paper.

Since $\mu$-law $L_1$ loss and $Tan\_L_1$ loss emphasize on different kinds of extreme values and thus have different performance on PSNR and PSNR-$\mu$, the proposed loss gives a good balance between PSNR and PSNR-$\mu$.


\section{Experiments and Results}
\label{sec:exp}

\subsection{Dataset}
Our dataset is provided by NTIRE 2022 HDR challenge\cite{perezpellitero22}, containing 1494 input LDR triplets with their respective HDR ground-truth for training, 60 triplets for validation, and 201 triplets for testing.

\begin{table*}[ht]
  \scalebox{2}{
  \tiny
  \centering
    \begin{tabular}{l c c c c c c c c}
      \toprule
      Exp. & MS & PDUB & share & align & conv-up & PSNR(db) & PSNR-$\mu$(db) & MAccs(G) \\
      \midrule
      1 &  &  & Y &  &  & 38.23	& 36.48 & 152.11 \\
      \midrule
      2 & $\checkmark$ &  & Y &  &  & 38.54 & 36.70 & 137.66\\
      3 & $\checkmark$ & $\checkmark$ & Y &  &  & 38.67 & 36.87 & 121.03 \\
      4 & $\checkmark$ & $\checkmark$ & Y & full  &  & 38.24 & 36.79 & 146.28\\
      5 & $\checkmark$ & $\checkmark$ & N &  &  & 38.60 & 36.79 & 121.03\\
      6* & $\checkmark$ & $\checkmark$ & N & full &  & 38.34 & 36.92 & 146.28 \\
      7 & $\checkmark$ & $\checkmark$ & N & half &  & 38.54 & 36.89 & 126.43 \\
      \midrule
      8* & $\checkmark$ & $\checkmark$ & N & half & $\checkmark$ & 38.74 & 37.02 & 198.38\\
      \bottomrule
    \end{tabular}
  }
  \caption{Ablation studies of effectiveness of network modules based on $\mu$-law $L_1$ loss.
    Exp. 6 is our lightweight model, and this is the model version in the listed ranking results for Track 2 in \cref{tab:comp_track2}. 
    In exp. 7, we half both the feature map resolution of attention and alignment module. 
    Exp.8 is our standard model, and this is the model version in the listed ranking results for Track 1 in \cref{tab:comp_track1}.}
  \label{tab:ablation}
\end{table*}

\begin{table}[t]
  \centering
  \resizebox{.48\textwidth}{!}{
    \begin{tabular}{l c c c c}
      \toprule
      \#PDUBs & PSNR(db) & PSNR-$\mu$(db) & MAccs(G) \\
      \midrule
      2 & 38.45 & 36.77 & 128.25\\
      3 & 38.34 & 36.92 & 146.28\\
      4 & 38.74 & 36.95 & 164.31\\
      5 & 38.77 & 36.97 & 182.34\\
      \bottomrule
    \end{tabular}
  }
  \caption{Comparison on number of PDUBs. Experiments are conducted based on lightweight model with $\mu$-law $L_1$ loss.}
  \label{tab:abl_pb}
\end{table}

\begin{table}[t]
  \centering
  \resizebox{.48\textwidth}{!}{
  \begin{tabular}{l c c c}
    \toprule
    Loss & PSNR(db) & PSNR-$\mu$(db)\\
    \midrule
    $L_1$ & 38.30 & 36.40\\
    $Tan\_L_1$ & 38.75 & 36.91 \\
    $\mu$-$law$ $L_1$ & 38.34 & 36.92\\
    $Tan\_L_1$ + $\mu$-$law$ $L_1$ & 38.66 & 36.92\\
    $\mu$-$law$ $L_1$ + $half^*$ & 38.54 & 36.89\\
    $Tan\_L_1$ + $\mu$-$law$ $L_1$ + $half^*$ & 39.00 & 36.96\\
    \bottomrule
  \end{tabular}
  }
  \caption{Comparison on losses. Experiments are conducted based on lightweight model.
   With experiments noted by $*$, we downsample the feature map resolution of attention module and alignment module.}
  \label{tab:abl_loss}
\end{table}

\begin{figure*}[th]
  \centering
  \begin{subfigure}{0.16\textwidth}
    \includegraphics[width=\textwidth]{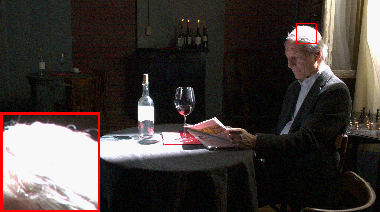}
    \vfill
    \includegraphics[width=\textwidth]{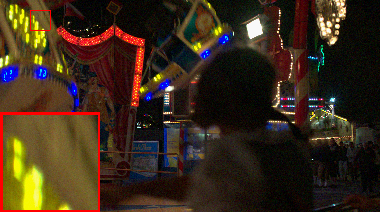}
    \vfill
    \includegraphics[width=\textwidth]{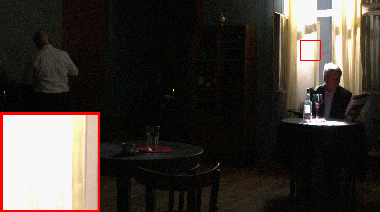}
    \caption{}
    \label{fig:d_1}
  \end{subfigure}
  \hfill
  \begin{subfigure}{0.16\textwidth}
    \includegraphics[width=\textwidth]{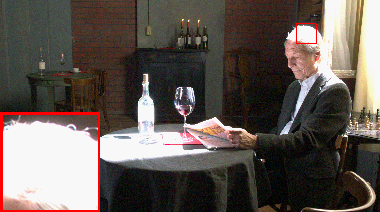}
    \vfill
    \includegraphics[width=\textwidth]{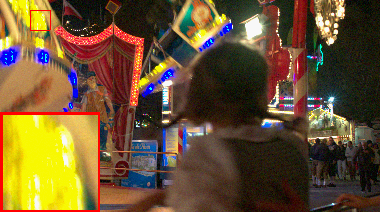}
    \vfill
    \includegraphics[width=\textwidth]{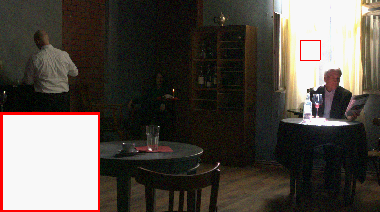}
    \caption{}
    \label{fig:d_2}
  \end{subfigure}
  \hfill
  \begin{subfigure}{0.16\textwidth}
    \includegraphics[width=\textwidth]{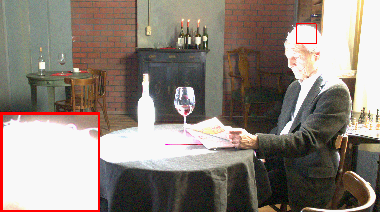}
    \vfill
    \includegraphics[width=\textwidth]{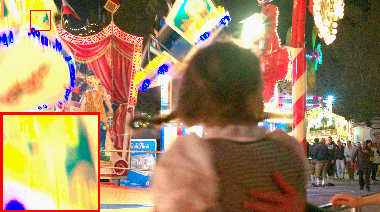}
    \vfill
    \includegraphics[width=\textwidth]{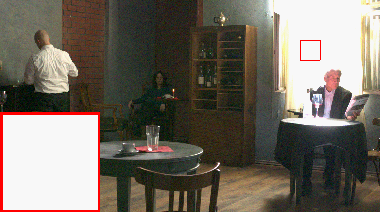}
    \caption{}
    \label{fig:d_3}
  \end{subfigure}
  \hfill
  \begin{subfigure}{0.16\textwidth}
    \includegraphics[width=\textwidth]{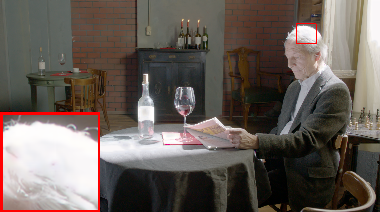}
    \vfill
    \includegraphics[width=\textwidth]{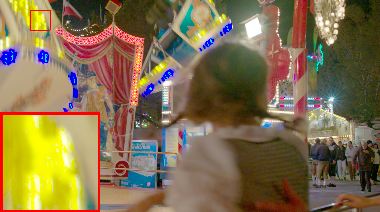}
    \vfill
    \includegraphics[width=\textwidth]{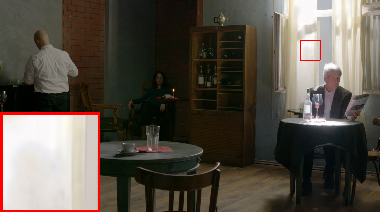}
    \caption{}
    \label{fig:d_6}
  \end{subfigure}
  \hfill
  \begin{subfigure}{0.16\textwidth}
    \includegraphics[width=\textwidth]{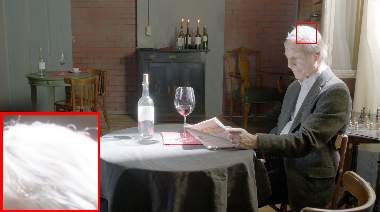}
    \vfill
    \includegraphics[width=\textwidth]{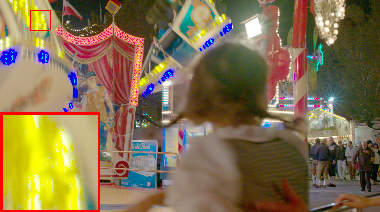}
    \vfill
    \includegraphics[width=\textwidth]{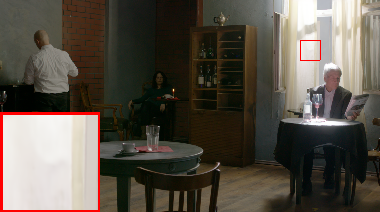}
    \caption{}
    \label{fig:d_5}
  \end{subfigure}
  \hfill
  \begin{subfigure}{0.16\textwidth}
    \includegraphics[width=\textwidth]{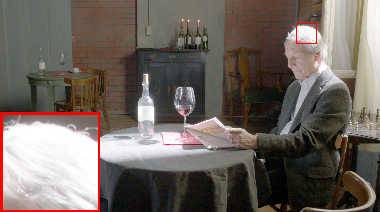}
    \vfill
    \includegraphics[width=\textwidth]{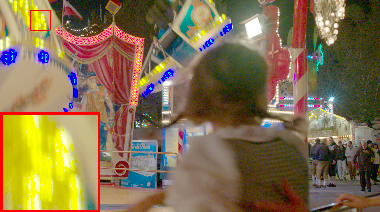}
    \vfill
    \includegraphics[width=\textwidth]{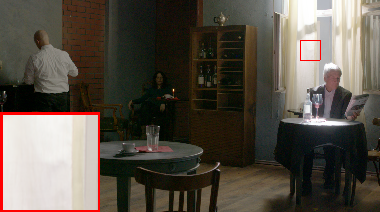}
    \caption{}
    \label{fig:d_4}
  \end{subfigure}
  \caption{Qualitative comparison of our EAPNet with SOTA method AHDR.
  \cref{fig:d_1}, \cref{fig:d_2} and \cref{fig:d_3} are short, medium, and long exposed LDRs respectively.
  The HDR images restored by AHDR are shown in \cref{fig:d_6},
  and the HDR images restored by our standard EAPNet are shown in \cref{fig:d_5},
  and the HDR ground-truth images are shown in \cref{fig:d_4}.}
  \label{fig:details}
  
\end{figure*}


\subsection{Implementation Details}
We set the number of PDUBs as 4, downsample the alignment feature map resolution, and apply transposed convolution in PDUB for our standard EAPNet model. 
While for the lightweight model, we cut one PDUB, keep full resolution for alignment module, and use bilinear upsampling to save computation cost and speed up.

During the training stage, we crop the input LDRs and ground-truth to 256x256-sized patches with an overlap of 128.
The network is trained from scratch, and optimized by an Adam optimizer \cite{kingma2014method}.
We train the model for 10 million iterations with a batch size of 32.
The initial learning rate is set as 8e-4, and decayed every 100k iterations by a factor of 0.75.
We use gradual warmup strategy \cite{goyal2017accurate} for the first 1000 steps to overcome early optimization difficulties.
All models are built on the PyTorch framework, and trained with 4 NVIDAI V100 GPUs, costing about 7 days.

Temporal ensemble \cite{laine2016temporal} is applied to generate an exponential moving average (EMA) model of the trained model.
We update the EMA model every 5k steps with a decay rate of 0.5.
The final EMA model is used to generate predictions for testing.

During the validation and testing stage, we pad the input images from 1900x1060 to 1920x1080, and crop to the original size after model inference.
Inferencing is conducted on NVIDAI V100 GPU.
It takes 0.24 second per image with 198.38G MAccs for our standard model, and 0.15 second per image with 146.28G MAccs for our lightweight model specifically.

\subsection{Results}
We participated in the NTIRE 2022 HDR Challenge and won the second place in both track 1 and track 2.
The comparisons of our method with other top methods and the challenge baseline are listed in \cref{tab:comp_track1} and \cref{tab:comp_track2}.

We also compare our method with some state-of-the-art methods, including AHDR \cite{yan2019attention} and ADNet \cite{ADNet}.
The quantitative results are provided in \cref{tab:sota}.
In terms of fidelity, our standard EAPNet exceeds AHDR provided by NTIRE 2022 HDR organizers by 0.16db in PSNR-$\mu$ and 0.4db in PSNR, with about 15$\times$ compression in MAccs.
And in terms of complexity, our lightweight EAPNet achieves about 20$\times$ compression in MAccs,
with PSNR maintained the same and a slight improvement in PSNR-$\mu$.

We present some reconstructed HDR examples in \cref{fig:details} produced from LDRs by EAPNet.
Details show that long-exposed and short-exposed areas are recovered from LDRs, and fine details are restored in the reconstructed HDRs.

\begin{table}[t]
  \centering
  \resizebox{.48\textwidth}{!}{
    \begin{tabular}{l c c c c c}
      \toprule
      Team & PSNR(db) & PSNR-$\mu$(db) & Runtime(s) & MAccs(G) & Params.(k) \\
      \midrule
      ALONG & 39.417 & 37.424 & 0.324 & 198.47 & 489.01\\
      Ours & 38.607 & 37.252 & 0.185 & 198.38 & 576.23\\
      XPixel-UM & 38.015 & 37.209 & 0.276 & 199.88 & 1013.25\\
      AdeTeam & 39.001 & 37.163 & 0.134 & 156.12 & 188.99\\
      CZCV & 37.388 & 36.972 & 0.431 & 193.93 & 633.69\\
      \bottomrule
    \end{tabular}
  }
  \caption{Results of NTIRE 2022 HDR Challenge Track 1 top 5 methods on online testset\cite{perezpellitero22}. Standard EAPNet is used in Track 1.}
  \label{tab:comp_track1}
\end{table}

\begin{table}[t]
  \centering
  \resizebox{.48\textwidth}{!}{
    \begin{tabular}{l c c c c c}
      \toprule
      Team & PSNR(db) & PSNR-$\mu$(db) & Runtime(s) & MAccs(G) & Params.(k)\\
      \midrule
      ALONG & 38.843 & 37.033 & 0.183 & 74.02 & 177.41\\
      Ours & 38.766 & 37.100 & 0.155 & 146.28 & 393.73\\
      AdeTeam & 39.001 & 37.163 & 0.136 & 156.12 & 188.99\\
      BOE-IOT-AIBD & 38.150 & 37.225 & 0.473 & 756.78 & 1355.10\\
      MegHDR & 38.749 & 37.030 & 0.269 & 790.48 & 401.83\\
      \midrule
      Baseline(AHDR) & 37.597 & 37.021 & 0.760 & 2916.92 & 1141.28 \\
      \bottomrule
    \end{tabular}
  }
  \caption{Results of NTIRE 2022 HDR Challenge Track 2 top 5 methods on online testset\cite{perezpellitero22}. The result is produced by lightweight EAPNet.}
  \label{tab:comp_track2}
\end{table}

\begin{table}[t]
  \centering
  \resizebox{.48\textwidth}{!}{
  \begin{tabular}{l c c c}
    \toprule
     & PSNR(db) & \textbf{PSNR-$\mu$(db)} & MAccs(G) \\
    \midrule
    AHDR$^\dagger$ & 38.34 & 36.86 & 2916.92\\
    AHDR* & 38.98 & 36.82 & 2710.12\\
    ADNet & \textbf{39.34} & \textbf{37.21} & 6249.43\\
    \midrule
    Lightweight EAPNet & 38.34 & 36.92 & \textbf{146.28}\\
    Standard EAPNet & 38.74 & 37.02 & 198.38\\
    \bottomrule
  \end{tabular}
  }
  \caption{Comparison with state-of-the-art methods. AHDR$*$ is reproduced by authors, and AHDR$^\dagger$ is provided by NTIRE 2022 HDR organizers\cite{perezpellitero22}.}
  \label{tab:sota}
\end{table}

\subsection{Ablation Studies}
In this part, we present ablation studies to analysis the effectiveness of each part of our model. Experiment results are listed in \cref{tab:ablation}.

We replace the standard convolution in AHDR \cite{yan2019attention} with depthwise separable convolution \cite{related_mbnv1}, and then reduce the feature map resolution by half in the restoration network to reduce model complexity.
We take this model as our base model, with 152.11G MAccs, corresponding to exp. 1 in \cref{tab:ablation}.

\textbf{Multi-Scale Encoder.}
We replace the naive encoder block in the base model with our proposed Multi-Scale Encoder in exp.2 in \cref{tab:ablation}.
Specifically, we use a 2-scale encoder, and obtain 0.22db gain on PSNR-$\mu$ performance with about 15G compression in MAccs.

\textbf{DRDB V.S. PDUB.}
We conduct experiments to explore the performance of DRDB and PDUB in exp. 2 and 3 in \cref{tab:ablation}.
Results show that PDUB gets 0.17db higher on PSNR-$\mu$ and 0.07db higher on PSNR, with 16.6G MAccs lower than DRDB.

\textbf{Alignment module and weight sharing.}
Encoder and attention module weights are shared by default in AHDR. In our method, these two modules are set apart with no weight sharing, since short-exposed and long-exposed frames may have different types of features.
An additional alignment module helps to improve model performance by attending feature merging.
Exp. 4, 5 and 6 in \cref{tab:ablation} show that alignment module and no weight sharing strategy together help to improve model performance, and this is our lightweight model.

\textbf{Standard model.}
We further shrink the model by half the feature map resolution of alignment module,
and then we replace bilinear upsampling with transposed convolutional upsampling in exp.8, and obtain the final standard model.

\textbf{Number of PDUBs.}
As listed in \cref{tab:abl_pb}, experiments on the number of PDUBs show that the model achieves better performance as the number of PDUBs increases.
Considering model efficiency, we set the number of PDUBs to 4 for our standard model and 3 for our lightweight model.

\textbf{Loss.}
As listed in \cref{tab:abl_loss}, $\mu$-law $L_1$ yields slightly better PSNR-$\mu$, and with $Tan\_L_1$ we get higher PSNR.
The proposed loss achieves a balance between PSNR and PSNR-$\mu$.

\section{Conclusion}
  In this paper, we describe the solution EAPNet for the NTIRE 2022 HDR Track 1 and Track 2. We use depthwise separable convolution and downsample feature map resolution to improve efficiency. The multi-scale encoder module concatenates shallow and deep features which are strongly required by HDR restoration. We propose PDUB that can progressively recover HDR images with low computation cost. To deal with misalignment problems, we use a lightweight and highly efficient feature-alignment module rather than DCN. To summarize, our method achieves good performance and outperforms the state-of-the-art method proven by experiments.
{\small
\bibliographystyle{ieee_fullname}
\bibliography{egbib}
}

\end{document}